\documentclass{article}

\usepackage{PRIMEarxiv}

\usepackage[utf8]{inputenc} 
\usepackage[T1]{fontenc}    
\usepackage{hyperref}       
\usepackage{url}            
\usepackage{booktabs}       
\usepackage{amsfonts}       
\usepackage{nicefrac}       
\usepackage{microtype}      
\usepackage{lipsum}
\usepackage{fancyhdr}       
\usepackage{graphicx}       
\usepackage{geometry,amsmath,amsthm,amssymb}
\usepackage{subcaption} 
\usepackage{array}
\usepackage{caption}
\usepackage{float}
\usepackage{microtype}
\usepackage{tabularx}
\usepackage{adjustbox}
\usepackage{threeparttable}
\usepackage{multirow}
\usepackage{xcolor}
\usepackage{url}
\usepackage{enumitem}

\title{
OG-HFYOLO :Orientation gradient guidance and heterogeneous feature fusion for deformation table cell instance segmentation
\thanks{\textit{{Corresponding author}}}.
}

\author{
  Long Liu \\
  Nanchang Hangkong University \\
  \texttt{2304081200008@stu.nchu.edu.cn} \\
   \And
  Cihui Yang$^*$ \\
  Nanchang Hangkong University \\
  \texttt{yangcihui@nchu.edu.cn} \\
}

\begin{document}
\maketitle

\begin{abstract}
  Table structure recognition is a key task in document analysis.However, geometric deformations in deformed tables weaken the correlation between content and structural information, thereby hindering downstream tasks' ability to extract accurate content. To address this challenge, we propose the OG-HFYOLO model for fine-grained cell coordinate localization. The model integrates a Gradient-Orientation-Aware Extractor to enhance edge detection and introduces Heterogeneous Kernel Cross Fusion module to boost multi-scale feature learning, improving feature expression accuracy. Combined with a Scale-aware Loss function for better scale feature adaptation during training and mask-driven non-maximal suppression replacing traditional bounding-box suppression post-processing, the model achieves refined feature representation and superior localization performance. We further propose a data generator to address dataset limitations for fine-grained deformation table cell localization and construct the large-scale Deformation Wired Table (DWTAL) dataset. Experiments demonstrate that OG-HFYOLO achieves superior segmentation accuracy compared to  all mainstream instance segmentation models on the DWTAL dataset. The dataset and the source code are open source: \url{https://github.com/justliulong/OGHFYOLO}.
\end{abstract}


\section{Introduction}
In the digital era of expanding information, tables serve as primary carriers of structured data, commonly conveying critical information in financial reports, educational materials, and scientific experimental results. Concurrently, the widespread adoption of scanning, photography, and other technologies has increased the complexity of table electronic document scenarios. Particularly under varying illumination, angles, and environmental conditions, scanned or photographed table images may exhibit diverse deformations. The physical deformations, such as bending, perspective distortion, and folds, pose significant challenges for table structure recognition technologies.

Table structure recognition technology, which reconstructs row-column topology and semantic information from table images, has advanced considerably through deep learning approaches \cite{tsr}. While numerous end-to-end models exist and they can generate HTML or LaTeX table structure sequences directly from images \cite{pubtablenet,latex-seq}, their black-box nature makes the intermediate steps uncontrollable. Conversely, non-end-to-end strategies adopt a modular approach that decouples cell localization and structural inference, offering greater transparency in table structure recognition. However, one of the core task of such methods lies in precisely determining the spatial coordinates of table cells, which requires identifying the pixel-level positions of each cell within the image.

\begin{figure}[H]
  \centering
  \begin{subfigure}[b]{0.3\textwidth}
      \includegraphics[width=\textwidth]{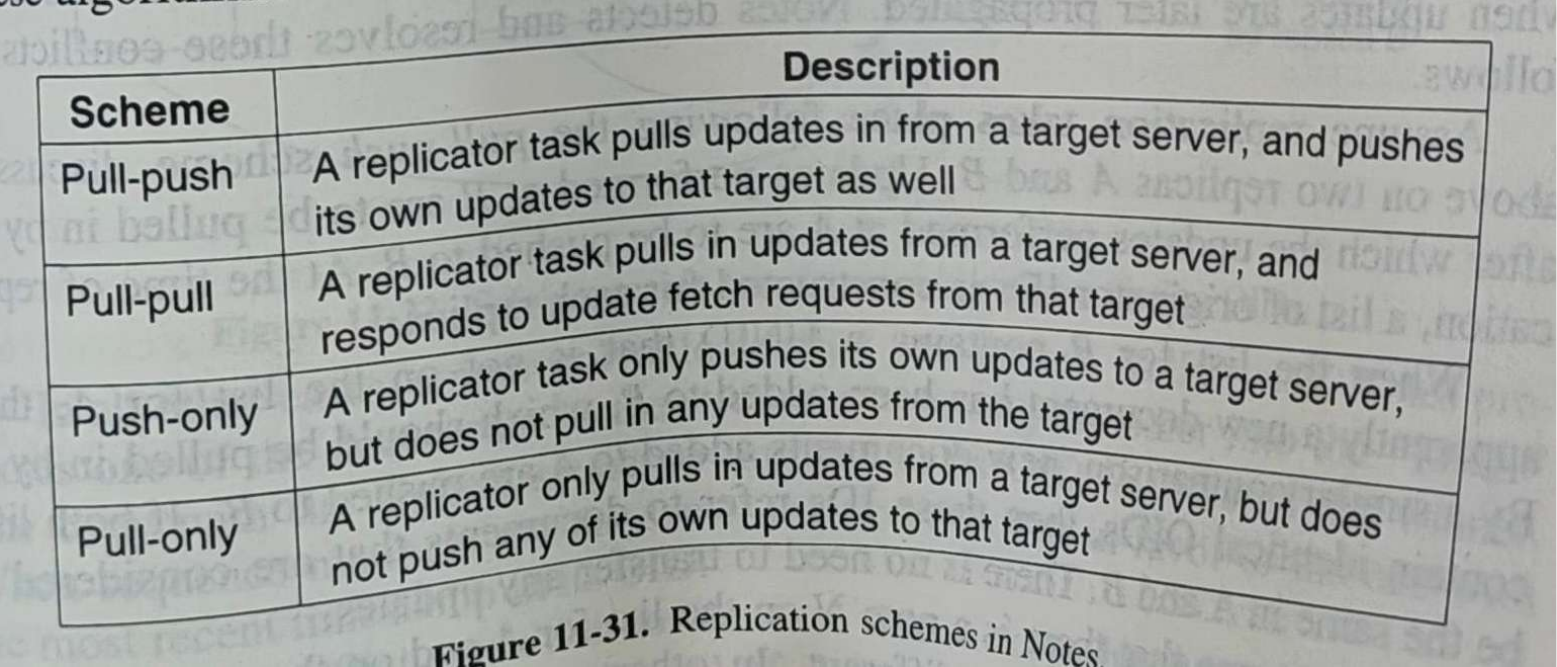}
      \caption{origin image}
      \label{fig:a}
  \end{subfigure}
  \begin{subfigure}[b]{0.3\textwidth}
      \includegraphics[width=\textwidth]{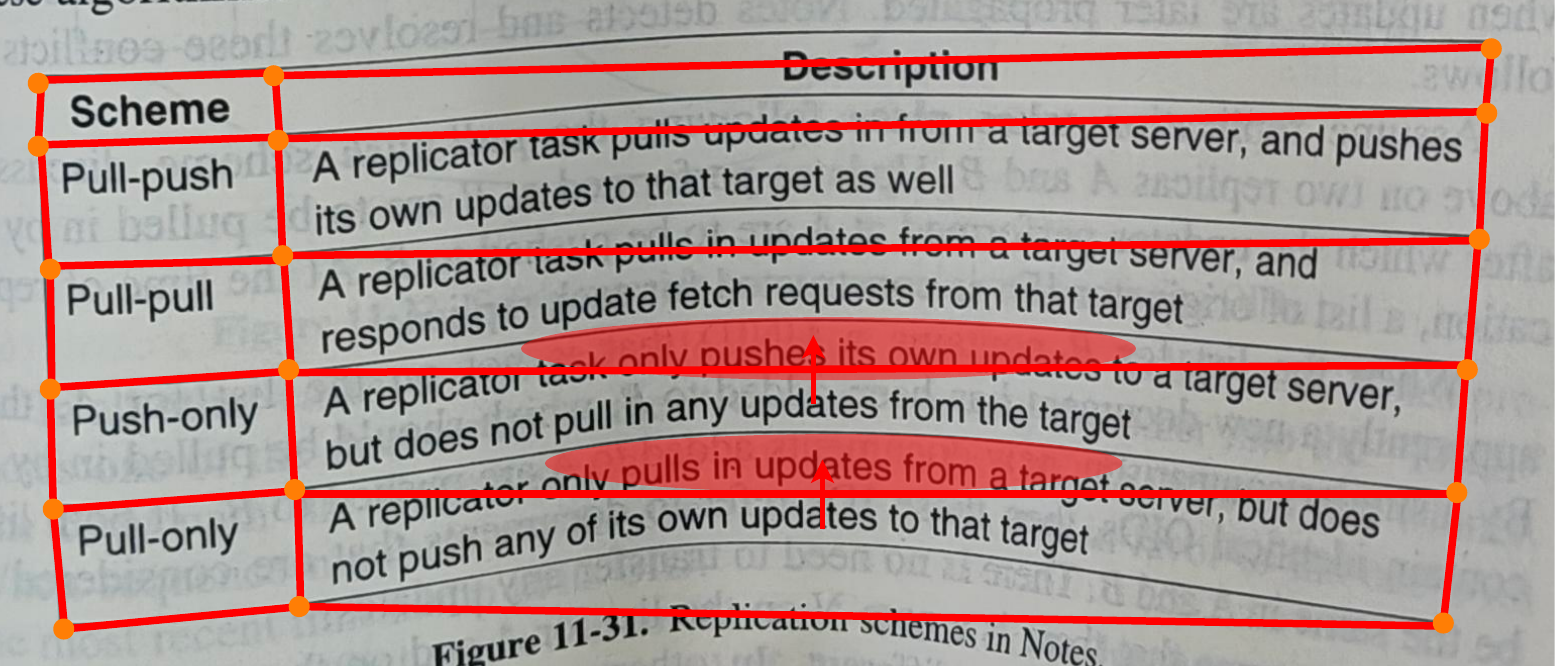}
      \caption{keypoint detection}
      \label{fig:b}
  \end{subfigure}
  \begin{subfigure}[b]{0.3\textwidth}
      \includegraphics[width=\textwidth]{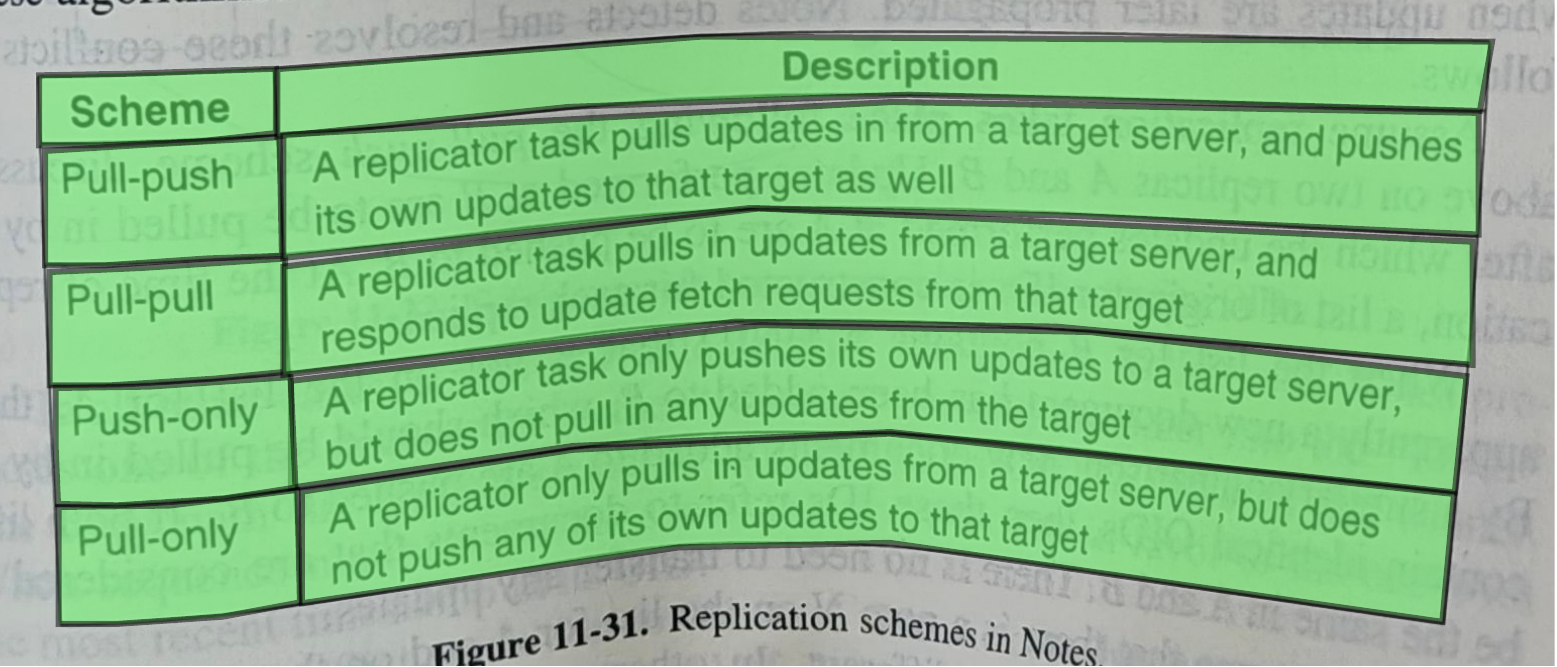}
      \caption{instance segmentation}
      \label{fig:c}
  \end{subfigure}
  \captionsetup{labelsep=colon} 
  \caption{Effect comparison of different methods: (\subref{fig:a}) shows a deformed table under natural scene, (\subref{fig:b}) demonstrates the detection results by contour corner-points, and (\subref{fig:c}) presents the instance segmentation-based contour. As shown, instance segmentation maintains complete content in deformed cells, providing richer information for downstream tasks like content restoration.}
\end{figure}

Current approaches to cell spatial coordinate localization can be broadly categorized into two main methods: One approach involves contour-based object detection, which can effectively identify cells with minor deformations. However, this method may omit critical cell content information with severe deformations, weakening the correlation between content and structure and complicating downstream tasks like content extraction. The other approach involves text box segmentation \cite{cascadetabnet,lpgma}, primarily used in wireless non-deformed tables but struggles with wired tables exhibiting geometric deformations. While keypoint detection (Figure \ref{fig:b}) partially addresses these challenges, it cannot fully resolve deformation-induced misalignment, leaving a critical gap in accurate coordinate acquisition for severely deformed wired cells. 

To bridge this gap, we propose an instance segmentation-based framework (Figure \ref{fig:c}), which achieves pixel-level spatial localization and provides comprehensive structural information to support downstream tasks.However, two major challenges arise when applying instance segmentation to spatial localization of cells in deformed tables: First, the dense arrangement of table cells and shared contour lines between adjacent cells complicate the extraction of boundary information. Second, merged cells result in significant variations in cell scales, such as overly wide or narrow cells, further complicating cell spatial localization. Instance segmentation aims to achieve pixel-level object separation and semantic parsing, reflecting the exploration of balancing accuracy and efficiency in computer vision. Early research primarily utilized two-stage frameworks. A representative example is Mask R-CNN \cite{mrcnn}, which achieved highly accurate mask predictions in complex scenes through region proposal generation and feature extraction. However, due to the inherent latency of multi-stage computational processes, such methods struggle to meet real-time application demands. With the emergence of single-stage methods, researchers abandoned the region proposal mechanism, instead directly predicting object locations and masks via intensive feature map analysis. Nevertheless, single-stage models still encounter significant challenges in handling complex object outlines and differentiating dense objects, requiring feature expression mechanism innovations to enhance accuracy.

Beyond the challenges of the dense arrangement of objects and large-scale variations, current datasets for deformed table recognition tasks exhibit certain limitations. As per existing literature, while there are a few large-scale open-source datasets such as WTW\cite{wtw} and TAL-OCR\cite{tabocr} that focus on deformed tables, their annotation granularity is insufficient to support the training of pixel-level segmentation models. To address this issue, we developed a data generator capable of creating new datasets from existing ones. Specifically, we selected mildly deformed tables from TAL-OCR and WTW, refined their labels into segmentation annotations, and used this annotated data with the data generator to produce a substantial amount of deformed table data, effectively supporting the training of instance segmentation models. Leveraging data from the open-source datasets TAL-OCR and WTW, along with 150 offline-collected and labeled images, we utilized the data generator to create a large-scale dataset. This dataset was split into two independent subsets based on their sources: DWTAL-s, comprising 8,765 simpler tables primarily from TAL-OCR, and DWTAL-l, consisting of 19,520 complex tables mainly expanded from WTW.

To address the challenges of dense objectives and scale variations in the derived dataset, we first introduced the Gradient Orientation-aware Extractor (GOE). This module incorporates image texture gradient direction and strength into the YOLO-based network as learning features, enabling more comprehensive capture of cell boundary information. Additionally, during the feature fusion stage, the model employs the Heterogeneous Kernel Cross Fusion(HKCF), which integrates the bottleneck structure \cite{neckb} and Heterogeneous Kernel Select Protocol (HKSP) \cite{hksp}. This enhances the fusion of features across scales, improving the model's detection performance under significant scale variations. Furthermore, we designed a Scale-aware loss function for segmentation tasks. This function adaptively assigns weights to  various scale objects, enhancing the model's robustness to scale variations. In the post-processing stage, traditional instance segmentation models typically employ bounding box-based non-maximal suppression (NMS). However, due to the dense arrangement and complex shapes of deformed cells, bounding boxes exhibit significant overlap. This causes boundary box-based NMS to incorrectly suppress correct objects. To address this issue, we introduced mask-based NMS, which directly performs suppression judgments using the intersection over union (IoU) of predicted masks. The main contributions of this work are summarized as follows:
\begin{itemize}
  \item To address the issue of object omission in densely arranged cells of deformed tables, we propose the Gradient Orientation-aware Extractor (GOE), which enhances the boundary discrimination of densely arranged cells. The GOE integrates two key cues: gradient intensity and orientation. During feature extraction, directional filters decompose the image gradient field, quantifying intensity distribution in edge regions and preventing the isotropic diffusion of conventional convolution kernels along edges. Additionally, an orientation-aware attention mechanism is incorporated. This mechanism encodes gradient directions as channel components, guiding the deep network layers to focus on discriminative regions at deformation boundaries. We also propose the Heterogeneous Kernel Cross Fusion (HKCF) module for the feature fusion stage. This module dynamically adapts feature representation for multi-scale objects via the Heterogeneous Kernel Select Protocol (HKSP), deploying asymmetric convolution kernels in parallel. By integrating cross-layer feature interaction mechanisms, the HKCF explicitly captures distinct spatial patterns of horizontally wide and vertically narrow cells.
  \item We further design a scale-aware loss function that adaptively assigns weights to objects based on their scales. This enhances the model's sensitivity to smaller deformed cells while reducing feature redundancy in larger cells. Furthermore, to prevent valid cells from being incorrectly suppressed during post-processing due to the complex shapes and dense arrangements of deformed cells, we replace conventional bounding box-based non-maximum suppression (NMS) with a mask-driven NMS post-processing operation.
  \item To address the scarcity of fine-grained annotated datasets for deformed tables, we designed a data generator that expands existing datasets. We first selected mildly deformed tables from existing datasets and converted their labels into finer-grained segmentation annotations. These curated samples were then processed by the generator to produce two datasets with varying difficulty levels and scales: DWTAL-s and DWTAL-l.
\end{itemize}

\section{Related Work}
\subsection{Spatial coordinate localization of table cells}
Table cell spatial coordinate localization, a critical upstream task in table structure recognition, is essential for subsequent table parsing and structure recognition. Researchers have proposed various methods to address cell localization. For example, Prasad et al. \cite{cascadetabnet} presented CascadeTabNet. This model treats table text boxes as objects and formulates text box detection as an instance segmentation task, extracting text box masks using Cascade Mask R-CNN. This marked the first application of instance segmentation algorithms in table structure recognition. However, CascadeTabNet's experiments were restricted to non-deformable tables, and the impact of empty cells on downstream tasks was not extensively explored. Following a similar concept, Qiao Liang et al. \cite{lpgma} developed the LGPMA model. This model integrates global and local information via a soft feature pyramid and uses instance segmentation for text box detection. Unlike CascadeTabNet, LGPMA includes a specialized search algorithm for empty cells.

To address spatial coordinate extraction in deformable tables, Cycle-CenterNet \cite{wtw} proposed a contour corner point-based object detection method that predicts four contour points by detecting cell centers. Although this method advances deformable table recognition, it has limitations. Severe boundary curvature increases the regression difficulty for corner points, and using coarse-grained object detection frameworks may result in the loss of critical cell information. 

The challenges brought by deformed tables urgently require finer-grained techniques to preserve the key information of cells in the table. Driven by this requirement, our work integrates instance segmentation technology into spatial coordinate localization for deformable tables. By leveraging the pixel-level precision of instance segmentation, our approach achieves more refined spatial coordinate extraction, effectively addressing the challenges of cell localization in deformable table structures.
\subsection{Existing dataset}
The advancement of table structure recognition has been driven by numerous open-source datasets. Early datasets, such as UNLV \cite{unlv} and ICDAR-2013 \cite{icdar2013}, were mainly designed to evaluate traditional methods. These datasets contain limited samples (typically fewer than 1,000 images) and lack spatial coordinate annotations for table cells. These limitations diminish their utility for modern deep learning-based modular table recognition pipelines. As deep learning requires increasingly larger datasets, researchers such as Xu Zhong and Minghao Li introduced large-scale datasets like PubTabNet \cite{pubtablenet} and TableBank \cite{tablebank}. However, these datasets focus on structural annotations via HTML or LaTeX sequences while overlooking spatial coordinate labeling. Similarly, large-scale datasets like FinTab \cite{fintabnet} and SciTSR \cite{scitsr} include cell coordinates and row-column relationships but primarily consist of data derived from structured digital documents (e.g., PDFs or LaTeX-exported documents). 

These datasets are highly standardized, making them unsuitable for deformable table recognition. Although ICDAR-2019 \cite{icdar2019} attempted to address this gap by introducing scanned archival documents, its limited scale (3600 images) and deformation types are still inadequate. The CamCap \cite{camcap} dataset, explicitly designed for rule-based algorithms, contains only 85 deformed tables, far too few to support data-driven deep learning models. Existing methods lacked robust solutions for tables in natural scenes with complex backgrounds and deformations until the WTW dataset \cite{wtw} was introduced. This benchmark provides annotations specifically designed for such challenging scenarios. However, WTW uses coarse-grained spatial coordinate annotations based on four contour corner points, which are ineffective for severe deformations. Similarly, the education-focused TAL-OCR \cite{tabocr} dataset, collected from real-world photography, contains moderate deformations but lacks fine-grained instance segmentation-level spatial annotations for cells. None of these datasets adequately address the challenges of recognizing heavily deformed table structures.

\begin{table}[ht]
  \centering
  \caption{Benchmark Datasets for Table Structure Recognition Datasets}
  \label{tab:datasets}
  \begin{tabular}{c|c|c|c}
  \hline
  \textbf{Dataset}       & \textbf{Year} & \textbf{table} & \textbf{Deformation}\\ 
  \hline
  UNLV\cite{unlv}                  & 2012         & 558   &     \\ 
  ICDAR-2013\cite{icdar2013}         & 2013       & 156    &       \\ 
  PubTabNet\cite{pubtablenet}        & 2020       & 568K &       \\ 
  TableBank\cite{tablebank}          & 2019       & 417K &       \\ 
  FinTab\cite{fintabnet}          & 2020       & 89K &       \\ 
  SciTSR\cite{scitsr}                & 2019       & 15K  &       \\ 
  ICDAR-2019\cite{icdar2019}        & 2019       & 3.6K  &       \\ 
  CamCap\cite{camcap}                & 2015       & 85    & $\surd $       \\ 
  WTW\cite{wtw}                     & 2021        & 14K  & $\surd $     \\
  TAL-OCR\cite{tabocr}               & 2021        & 18K   & $\surd $   \\  
  \hline
  \end{tabular}
\end{table}

\subsection{Improvement of instance segmentation in YOLO}
Compared to two-stage instance segmentation models, single-stage YOLO-based models demonstrate a balanced performance in both speed and accuracy for instance segmentation tasks, attracting extensive research efforts to enhance the YOLO framework. For example, YOLOMask \cite{yolomask} and PR-YOLO \cite{pryolo} integrated the CBAM module into YOLOv5 \cite{yolov5} to reduce background noise interference. Similarly, YOLO-SF \cite{yolosf} incorporated the CBAM module into YOLOv7 \cite{yolov7} to improve sensitivity to small object features. YOLO-CORE \cite{yolocore} proposed multi-stage constraints (polar coordinate distance loss and sector loss) for direct contour regression to enhance mask boundary precision. YUSEG \cite{yuseg} combined UNet with YOLOv4 \cite{yolov4} to address ambiguous segmentation in dense objects. TTIS-YOLO \cite{ttisyolo} improved instance segmentation accuracy in complex road scenes through multi-scale efficient cross-stage modules, bidirectional cross-scale connections, and dynamic gradient optimization. GHA-Inst \cite{ghainst} alleviated instance occlusion and background interference by refining YOLOv7’s feature fusion and output layers while introducing a Global Hybrid Attention (GHA) module to enhance critical feature retention.

While these advancements cater to diverse scenarios, no existing studies, to the best of our knowledge, have applied YOLO-based segmentation models to spatial coordinate localization of cells in deformable tables. This task presents significant challenges due to extreme variations in object scales, complex cell shapes, and dense arrangements of objectives.

\section{Data generator and derivative dataset}
This section begins by outlining the implementation of the data generator, followed by a brief presentation of the DWTAL dataset, which is derived from this generator.
\subsection{Data generator}
Deep learning is inherently data-driven, with dataset scale and quality being critical factors in determining model performance. To generate a sufficiently large and high-quality dataset, we propose a data generator that creates distorted images and corresponding labels from existing images and annotations. The distortions are carefully designed to simulate real-world deformation scenarios. Specifically, the generator incorporates two distortion techniques: wave warping and cylindrical warping. Specifically, the generator incorporates two distortion techniques: wave warping and cylindrical warping. Additionally, it adjusts image brightness according to original illumination to mimic natural lighting variations in real-world photography. The subsequent three subsections detail the implementation of these two distortion techniques and the brightness adjustment strategy.
\subsubsection{Wave distortion transformation}
Wave warping is an image distortion technique based on trigonometric functions that creates wave-like deformations. This technique simulates distortions caused by attaching images to flexible curved surfaces (e.g., corrugated pipes or folded fabrics), which involve sine or cosine function-based warping, as well as human-induced distortions. Such warping introduces severe deformations, posing significant challenges for table structure recognition. The specific trigonometric warping transformation is defined as follows in Equation \ref{eq:trigonometric}:
\begin{equation}
  \left\lbrack  
  \begin{array}{l} 
      x \\  y 
  \end{array}
  \right\rbrack  
  + 
  \left\lbrack  
  \begin{array}{l} 
      \mathrm{A} \cdot  \sin \left( \frac{2\pi y}{\omega }\right) \\  \mathrm{A} \cdot  \cos \left( \frac{2\pi x}{\omega }\right)  
  \end{array}
  \right\rbrack
  = 
  \left\lbrack  
  \begin{array}{l} 
      {x}^{\prime } \\  {y}^{\prime } 
  \end{array}
  \right\rbrack
\label{eq:trigonometric}
\end{equation}
\noindent where $x$ and $y$ denote the coordinates in the original image, while $x'$ and $y'$ represent the coordinates after distortion. The parameter $\omega$ corresponds to the wavelength, which controls the distortion period (the larger the value of $\omega$, the shorter the distortion period, resulting in more frequent image perturbations). The amplitude $A$ controls the distortion intensity, with higher values of $A$ leading to more severe deformations. Both $\omega$ and $A$ are adjustable parameters.
\subsubsection{Cylindrical distortion transformation}
Cylindrical warping distorts image content to simulate cylindrical surface deformations. This method mimics deformations caused by binding line compression in books or reports, photographs of cylindrical surfaces, or curved distortions from photographing a document held aloft. Such deformations are common in real-world scenarios and address the practical requirement for document upright orientation. Consequently, the warping transformation is applied exclusively along the document's vertical orientation (y-axis) while preserving the horizontal axis (x-axis). The cylindrical warping transformation is defined as follows in Equation \ref{eq:cylinder}:
\begin{equation}
  \left\lbrack  
  \begin{matrix} x \\  y \cdot  \cos \left( \frac{\mathrm{F} \cdot  \left( {x - \frac{W}{\mathrm{c}}}\right) }{\frac{W}{\mathrm{c}}}\right)  
  \end{matrix}\right\rbrack 
  = 
  \left\lbrack  
  \begin{array}{l} 
      {x}^{\prime } \\  {y}^{\prime } 
  \end{array}
  \right\rbrack
  \label{eq:cylinder}
\end{equation}
\noindent where $x$ and $y$ denote the original image coordinates, $x'$ and $y'$ represent the warped coordinates, $W$ is the image width, $F$ serves as the distortion factor (with higher values of $F$ resulting in more pronounced distortion), and $c$ defines the warping axis parameter. The param $\frac{W}{\mathrm{c}}$ specifies the central axis of cylindrical curvature. Larger values of $c$ shift the axis leftward, near the axis to undergo milder distortion, while those farther away experience stronger warping. Both $F$ and $c$ are adjustable parameters in the formula.
\subsubsection{Illumination adjustment mechanism}
Photographic images in real-world scenarios often exhibit shadows due to lighting angles, with shadow locations varying with illumination direction. Most shadows originate from camera equipment, with the darkest areas typically localized near one of the four image corners. We propose an illumination adjustment mechanism that first calculates overall image brightness using Equation \ref{eq:light}. This equation is derived from the standard-definition television standard ITU-R BT.601 \cite{itu601}:
\begin{equation}
  L = 0.2989 \times R + 0.587 \times G + 0.114 \times B
  \label{eq:light}
\end{equation}
\noindent where $R$,$G$,$B$ represent the luminance values of the red, green, and blue channels in the image background, respectively, while $L$ denotes the overall brightness (higher $L$ values indicate brighter images).

If the image brightness falls below a predefined threshold, no additional shadows are applied. Otherwise, a shadow center is randomly selected near one of the four image vertices, and per-pixel brightness values are recomputed as follows:
\begin{equation}
  I(x,y) = {eb} + \left( {{cb} - {eb}}\right)  \cdot  \frac{{d}_{center}(x,y)}{{d}_{max}}
  \label{eq:x-y-light}
\end{equation}
\noindent where $cb$ denotes the Center Brightness (peak shadow intensity, with values in [0,1] where higher values indicate greater brightness), $eb$ represents Edge Brightness (weakest shadow intensity, also within [0,1]), ${d}_{max}$ is the image diagonal length, and ${d}_{center}(x,y)$ measures the Euclidean distance between pixel $(x,y)$ and the shadow center. Both $cb$ and $eb$ serve as adjustable parameters.

The data generator integrates three operations (wave distortion, cylindrical warping, and illumination adjustment) through coordinated parameter randomization. For wave distortion, the amplitude $A$ is uniformly sampled from $[10, 50]$, while the wavelength $\omega$ is dynamically constrained to avoid unrealistic high-frequency distortions: ${\omega}{\text{min}}$ increases proportionally to $A$ via a scaling factor $s \in [1, 5]$, and $\omega$ is finally selected from $[{\omega}{\text{min}}, 800]$. For cylindrical warping, the curvature axis $c$ follows a truncated normal distribution $\mathcal{N}(2, 0.7^2)$ within $[1, 5]$, and the deformation factor $F$ is adaptively sampled from a range determined by $c$—smaller $c$ (near the image center) corresponds to milder distortions ($F \in [0.7, 0.85]$), while larger $c$ (offset from the center) reduces $F_{\text{min}}$ linearly to enforce physically plausible attenuation. For illumination adjustment, shadows are added only if the image brightness exceeds a threshold, with center brightness $cb \in [0.6, 0.9]$ and edge brightness $eb \in [0.1, 0.3]$ controlling the intensity gradient. These inter-parameter dependencies ensure that generated distortions mimic natural variations while avoiding implausible artifacts.

\subsection{Derived dataset DWTAL}
The data generator enables large-scale synthesis of deformed table data from limited mildly deformed tables. Although DWTAL datasets includes 150 collected samples, it primarily relies on two public natural scene datasets: TAL-OCR and WTW. First, mildly deformed tables from TAL-OCR and WTW are selected to generate fine-grained segmentation masks using corner coordinates. These masks are then iteratively processed by the data generator to produce diverse deformed table images. The two source datasets have distinct characteristics. TAL-OCR contains education-focused images with clear table structures and simplified backgrounds, resulting in relatively simpler derived data. To enhance diversity, collected samples are merged with TAL-OCR-derived data to form a compact dataset called DWTAL-s, which has a total of 8,765 images. In contrast, WTW has more complex backgrounds and richer content, enabling the creation of a larger, more challenging dataset named DWTAL-l, which has a total of 19,520 images.

\noindent \textbf{Dataset Splitting.} Both datasets follow identical partitioning protocols. To ensure uniform distribution of deformation types across training and test sets, 80\% of each dataset is randomly allocated for training and 20\% for testing. Finally, the DWTAL-s dataset has 7,012 images in the training set and 1,753 images in the test set, while the DWTAL-l dataset has 15,616 images in the training set and 3,904 images in the test set.

The derived datasets retain key characteristics from their parent collections, such as curved deformations, perspective distortions, multicolor backgrounds, surface irregularities, and illumination variations. A significant improvement is the inclusion of pixel-level instance segmentation annotations for table cells. However, all images contain only single-table instances. Furthermore, a dataset version with logical coordinate annotations has been publicly released to facilitate wider research applications.

\begin{figure}
  \centering
  \includegraphics[width=\linewidth]{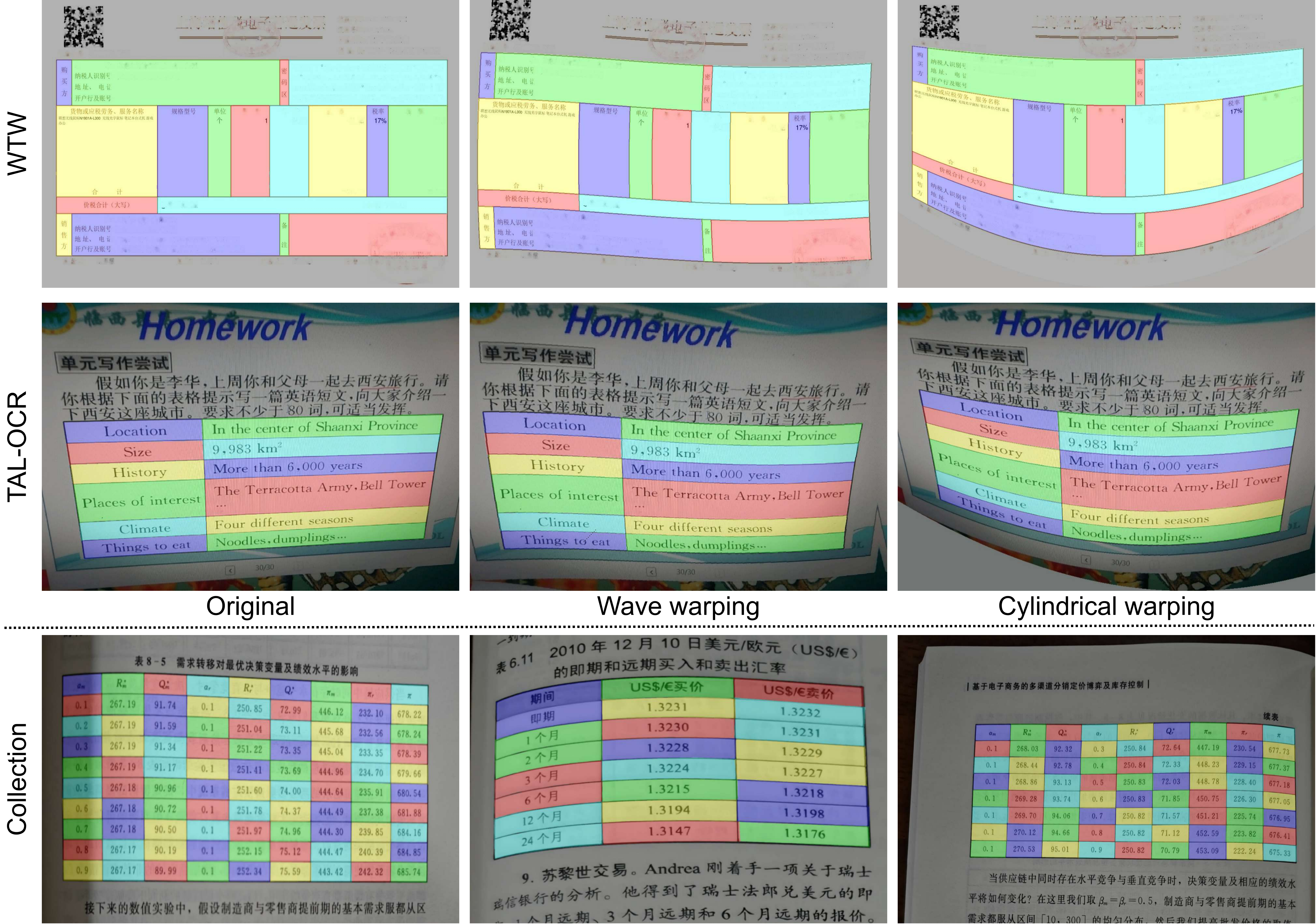}
  \caption{Dataset Annotation Examples: The source of each row's imagery differs: the first row originates from the WTW dataset, the second row is derived from the TAL-OCR dataset, and the third row comprises real-world images collected through offline acquisition.}
  \label{fig:dwtal}
\end{figure}

\section{Method}
\subsection{Overall architecture}
\begin{figure}[!htbp]
  \centering
  \includegraphics[width=0.9\linewidth]{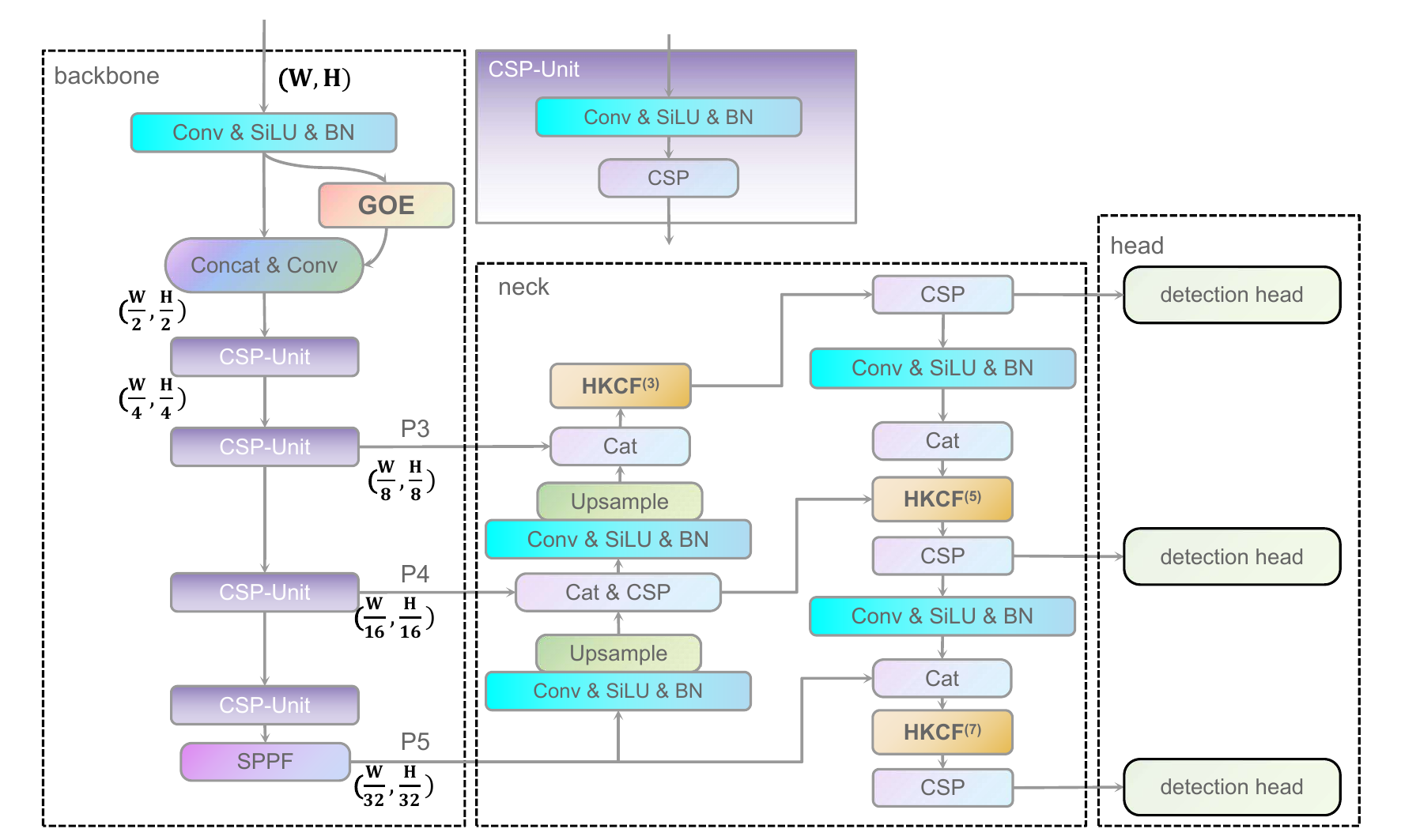}
  \caption{ The overview of the OG-HFYOLO.}
  \label{fig:og-hfyo}
\end{figure}

\autoref{fig:og-hfyo} illustrates the overall architecture of OG-HFYOLO, which adopts a YOLO framework and comprises three core components: the feature extraction backbone, the feature fusion neck, and the detection head. To enhance texture feature extraction, the model feeds the downsampled feature maps into the Gradient Orientation-aware Extractor (GOE) and concatenates them with the original input features along the channel dimension, generating fused feature representations enriched with texture information. The backbone employs CSP-Unit modules. Each module includes a $3 \times 3$ downsampling convolution, SiLU activation, batch normalization, and the Cross-Stage Partial Network (CSP) \cite{csp} utilized in YOLOv5. Through five downsampling operations, the backbone produces multi-scale feature maps {P3, P4, and P5}. 

During feature fusion, the extracted features are integrated via the FPN-PAN \cite{fpn-pan} pathway. Unlike standard YOLO implementations, our model incorporates a Heterogeneous Kernel Cross Fusion (HKCF) module after skip connections to enhance cross-scale feature interactions. The fused features are then refined via CSP blocks before being passed to the detection head. The fused features are then refined via CSP blocks before being passed to the detection head. The detection head retains YOLOv5's anchor-based \cite{anchorbase} design, performing classification and bounding box regression using predefined anchor boxes, with Non-Maximum Suppression \cite{nms} (NMS) filtering redundant detections.

\subsection{Gradient Orientation-aware Extractor}
We identify object density as the primary challenge in spatial coordinate localization for deformed table cells. Inaccurate contour extraction by the model may lead to the merging of adjacent small cells into erroneously detected large ones. This aggregation progressively obscures scale variations during feature fusion, ultimately reducing prediction accuracy. 

\begin{figure}[htb]
  \centering
  \includegraphics[width=0.6\linewidth]{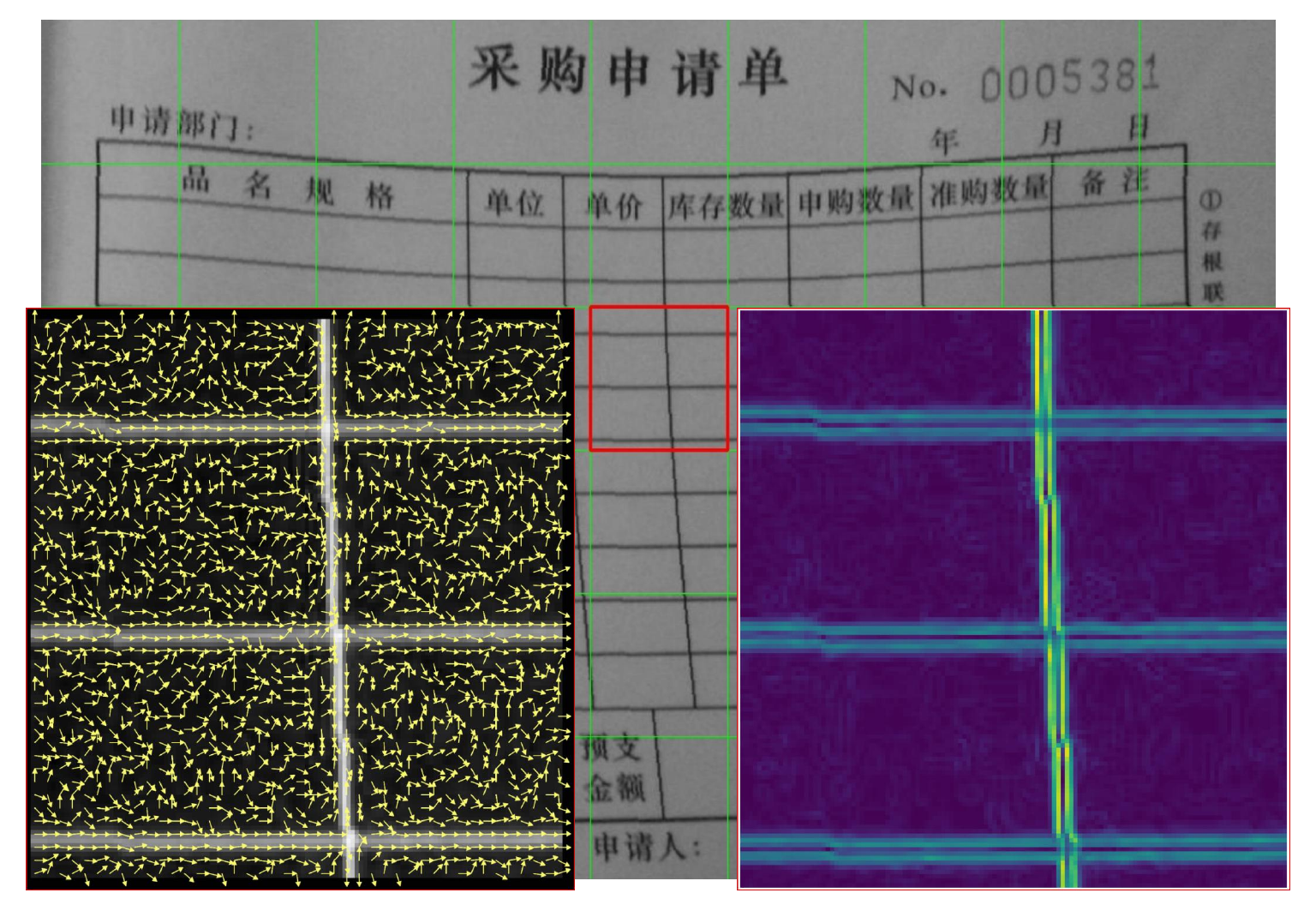}
  \caption{HOG feature extraction core schematic diagram: the left inset demonstrates gradient orientation features within the red grid, while the right(Heatmap: The brighter the area, the higher the intensity) inset illustrates gradient magnitude features within the same grid. These two key features allow for comprehensive features of edge texture information.}
  \label{fig:hog}
\end{figure}

The instance segmentation accuracy in densely packed scenes is limited by the model's ability to perceive contour details. The Histogram of Oriented Gradients (HOG) \cite{hog} is designed to detect complex human contours, positing that fine gradients and precise orientation binning are crucial for capturing intricate outlines. Its core principle provides significant insights for dense and complex object detection: as shown in Figure \ref{fig:hog}, HOG divides the image into multiple grids and decouples the gradient features within each grid into gradient direction and gradient intensity. It then statistically aggregates the gradient intensities in each direction to form a histogram. The core idea is to simultaneously capture the magnitude information of the gradients and the directional priors, thereby enhancing the geometric discriminability of the contours. 

The Gradient Orientation-aware Extractor(GOE) operates on this principle, enabling the model to learn both gradient intensity and gradient direction features of contour details. This effectively enhances its recognition capability for objects with complex contours and dense arrangements.

\begin{figure}[htb]
  \centering
  \includegraphics[width=\linewidth]{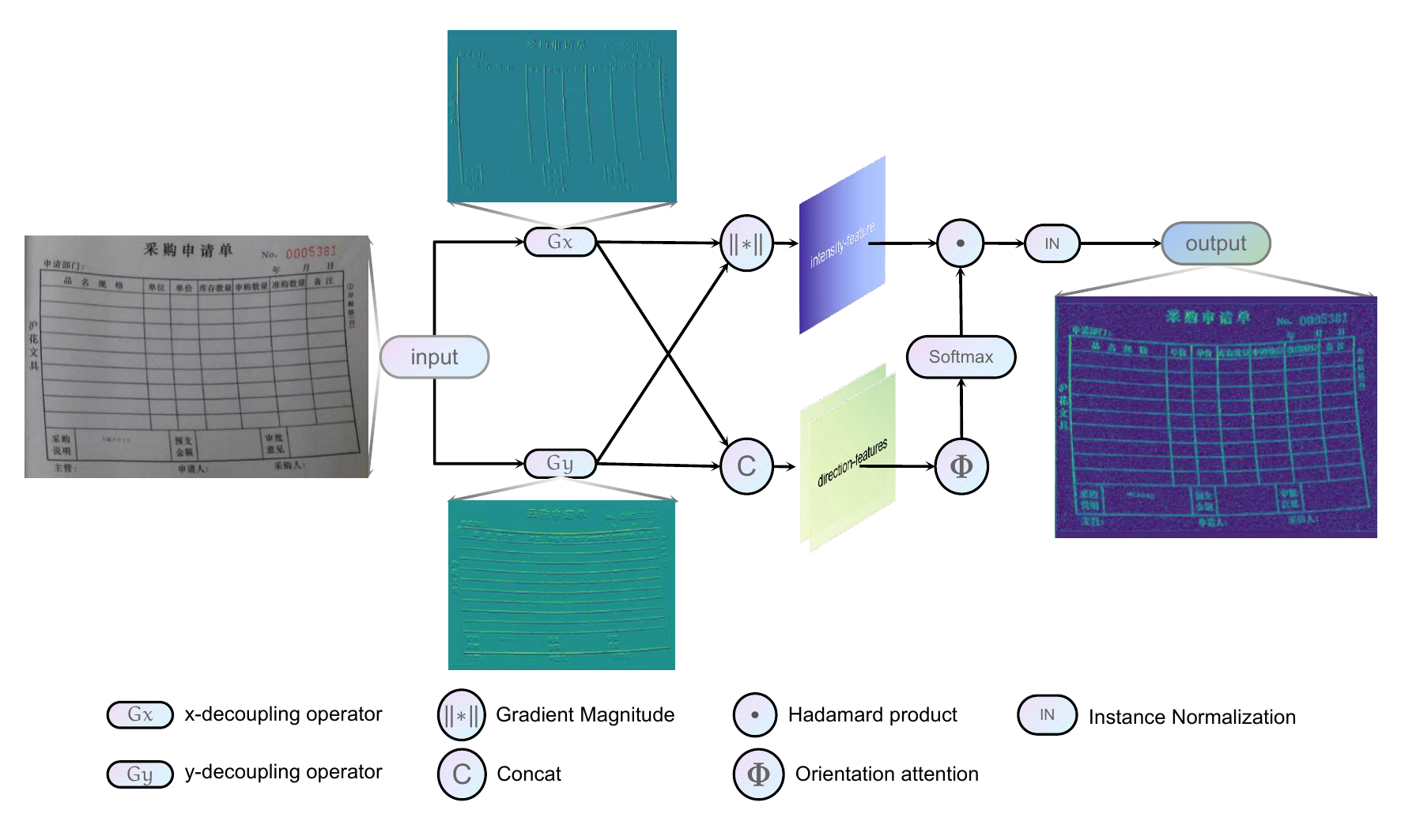}
  \caption{The overall structure of GOE: Feature extraction and combination of gradient direction and gradient magnitude through decoupling and recombination.}
  \label{fig:goe}
\end{figure}

\autoref{fig:goe} illustrates the internal architecture of the proposed Gradient Orientation-aware Extractor (GOE). The module takes a feature map encoding texture information as input. The GOE first uses distinct decoupling operators (Equation \ref{eq:gx_gy}) to decompose the input feature map $\mathbf{I}$ into a horizontal gradient orientation feature map ${G}_{x}$ and a vertical gradient orientation feature map ${G}_{y}$. 

\begin{equation}
  {G}_{x} , {G}_{y} = \mathcal{G}_{x} (\mathbf{I})  , \mathcal{G}_{y} (\mathbf{I})
  \label{eq:gx_gy}
\end{equation}

To preserve the directional priors of the decoupling operators $\mathcal{G}_{x}$ and $\mathcal{G}_{y}$, their weights are initialized as conventional edge operators. During training, convolution kernel parameter constraints are relaxed. This allows the network to adaptively adjust gradient response weights according to task requirements (e.g., deformation edge enhancement, noise suppression), thus overcoming the inherent geometric limitations of standard edge operators.

After obtaining the decoupled horizontal gradient orientation feature map ${G}_{x}$ and vertical gradient orientation feature map ${G}_{y}$, the module employs distinct strategies to integrate gradient features from both orientations: on one hand, the GOE calculates gradient magnitude by Equation \ref{eq:gm} to generate the gradient intensity feature map $I_{GM}$; on the other hand, the GOE performs channel-wise concatenation according to Equation \ref{eq:gd} to derive the gradient orientation feature map $I_{GD}$.
\begin{equation}
  I_{GM} = \sqrt{{G}_{x}^2 + {G}_{y}^2}
  \label{eq:gm}
\end{equation}

\begin{equation}
  I_{GD} = Cat({G}_{x} , {G}_{y})
  \label{eq:gd}
\end{equation}

The GOE module ultimately aggregates the gradient intensity feature map $I_{GM}$ and gradient direction feature map $I_{GD}$ according to Equation \ref{eq:goe}. It first applies orientation attention $\Phi$ to channel-encoded $I_{GD}$, mapping gradient information from distinct orientations into separate channels. The encoded features are then normalized via Softmax\cite{softmax}. Subsequently, the gradient intensity feature map $I_{GM}$ is weighted through the Hadamard product with the channel-encoded gradient direction feature map $I_{GD}$. Since gradients in each orientation are mutually independent, instance normalization (IN)\cite{IN} is adopted to stabilize model training, ultimately yielding the feature output $I_{o}$ that simultaneously encapsulates gradient orientation and intensity feature.
\begin{equation}
  I_{o} = IN(Softmax(\Phi(I_{GD})) \odot I_{GM})
  \label{eq:goe}
\end{equation}
\noindent \textbf{Orientation Attention $\Phi$:} The orientation attention mechanism $\Phi$ in the Gradient Orientation-aware Extractor (GOE) module is designed to encode gradient directionality through a learnable convolutional operation, combining geometric priors with adaptive feature learning. Inspired by the orientation binning strategy in Histogram of Oriented Gradients (HOG), we initialize the convolutional kernels using directional basis vectors derived from polar coordinate transformations. Specifically, the continuous angular space $[0, \pi)$ is uniformly divided into $n$ discrete bins, with the central angle of the $i$-th bin defined as $\theta_i = \frac{i\pi}{n}$. For each angle $\theta_i$, the corresponding Cartesian unit vector is computed as:
\begin{equation}
  \mathbf{e}_i = \begin{bmatrix}
    \cos\theta_i \\ 
    \sin\theta_i
    \end{bmatrix} \in \mathbb{R}^2, i=0,1,2,\ldots,n-1.
\end{equation}
explicitly representing gradient orientation in orthogonal coordinates. These vectors are structured into a parameter matrix $W \in \mathbb{R}^{n \times 2 \times 1 \times 1}$, where each row corresponds to a directional basis:  
\begin{equation}
  W = \begin{bmatrix}
    \cos\theta_0 & \sin\theta_0 \\
    \cos\theta_1 & \sin\theta_1 \\
    \vdots & \vdots \\
    \cos\theta_{n-1} & \sin\theta_{n-1}
    \end{bmatrix} \otimes \mathbf{I}_{1\times1},
\end{equation}
with $\otimes$ denoting the outer product expansion to match the 4D tensor structure of convolutional kernels. This matrix serves as the initial weights of a $1\times1$ convolution layer, mapping the input gradient fields (horizontal and vertical components) to $n$ orientation-specific feature channels. During training, the kernel weights are dynamically optimized through backpropagation, allowing the network to suppress noisy orientations while amplifying discriminative directional patterns. Unlike static HOG descriptors, this learnable mechanism adapts to task-specific requirements—for example, enhancing deformation boundaries in densely packed table cells. The final output combines orientation-aware features with the original texture representations through channel-wise concatenation, providing a geometrically grounded yet flexible foundation for downstream segmentation tasks.  

\begin{figure}[!htbp]
  \centering
  \includegraphics[width=\linewidth]{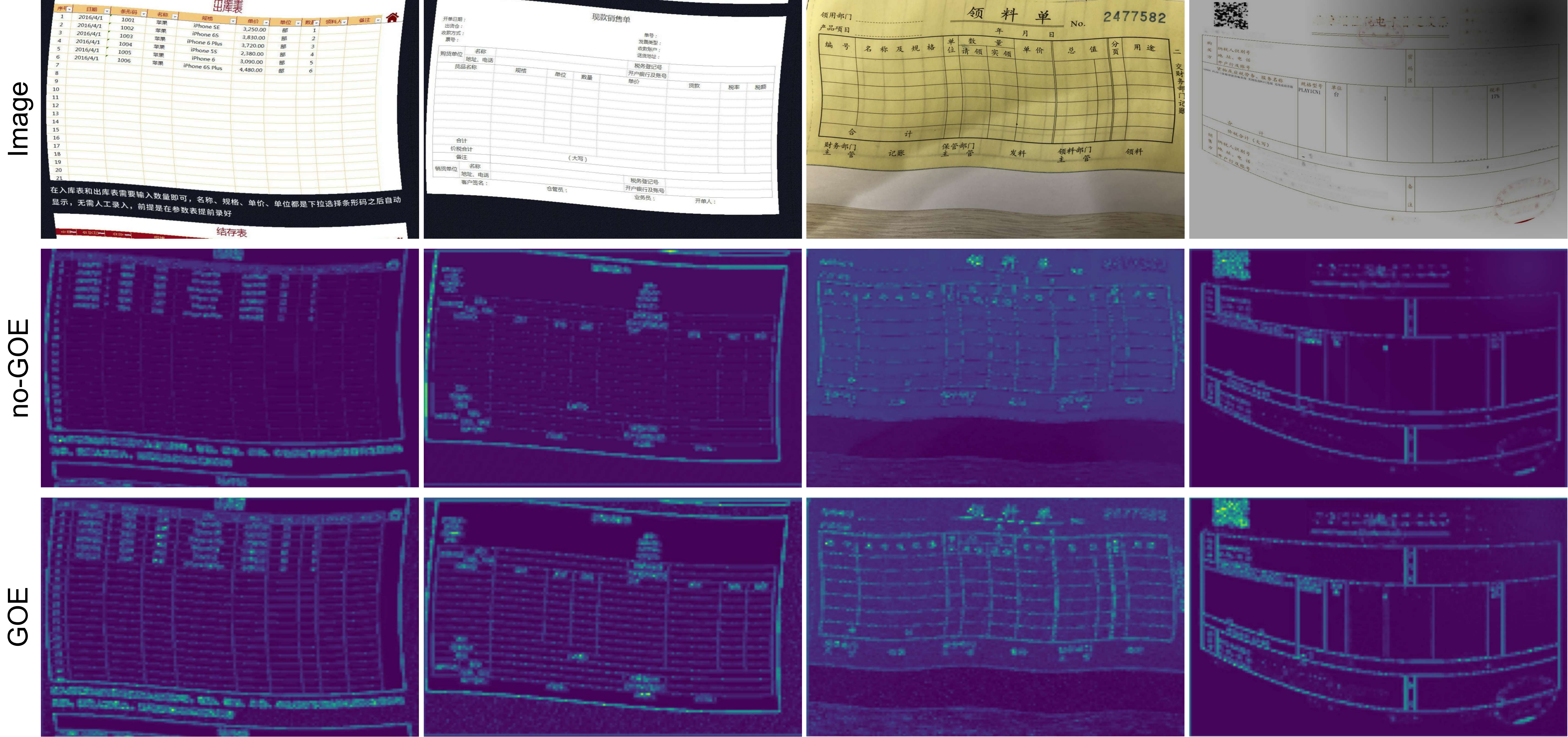}
  \caption{GOE module effect demonstration: The second line shows the effect of the second downsampling feature map of the model in the backbone without adding the GOE module, and the third line shows the effect of the second downsampling feature map of the model in the backbone with adding the GOE module.}
  \label{fig:show-goe}
\end{figure}

The hierarchical feature learning mechanism in convolutional neural networks shows that shallow, high-resolution feature maps are better at capturing low-level visual features such as edges and textures. In contrast, deeper layers focus more on semantic abstraction and global structure modeling \cite{vcnn}. Based on this, we integrate the Gradient Orientation Extractor (GOE) after the first convolutional downsampling layer in the backbone network. This leverages the high spatial resolution of shallow layers to precisely capture directional gradient features. Figure \ref{fig:show-goe} demonstrates the effectiveness of the GOE module in early feature extraction from the second downsampled feature map. By incorporating the proposed extractor, the network introduces geometric priors in the early stages for images with chromatic interference and blurred cell contours. This establishes a fine-grained foundation for subsequent cross-layer feature fusion.

\subsection{Heterogeneous Kernel Cross Fusion}
\begin{figure}[!htbp]
  \centering
  \includegraphics[width=\linewidth]{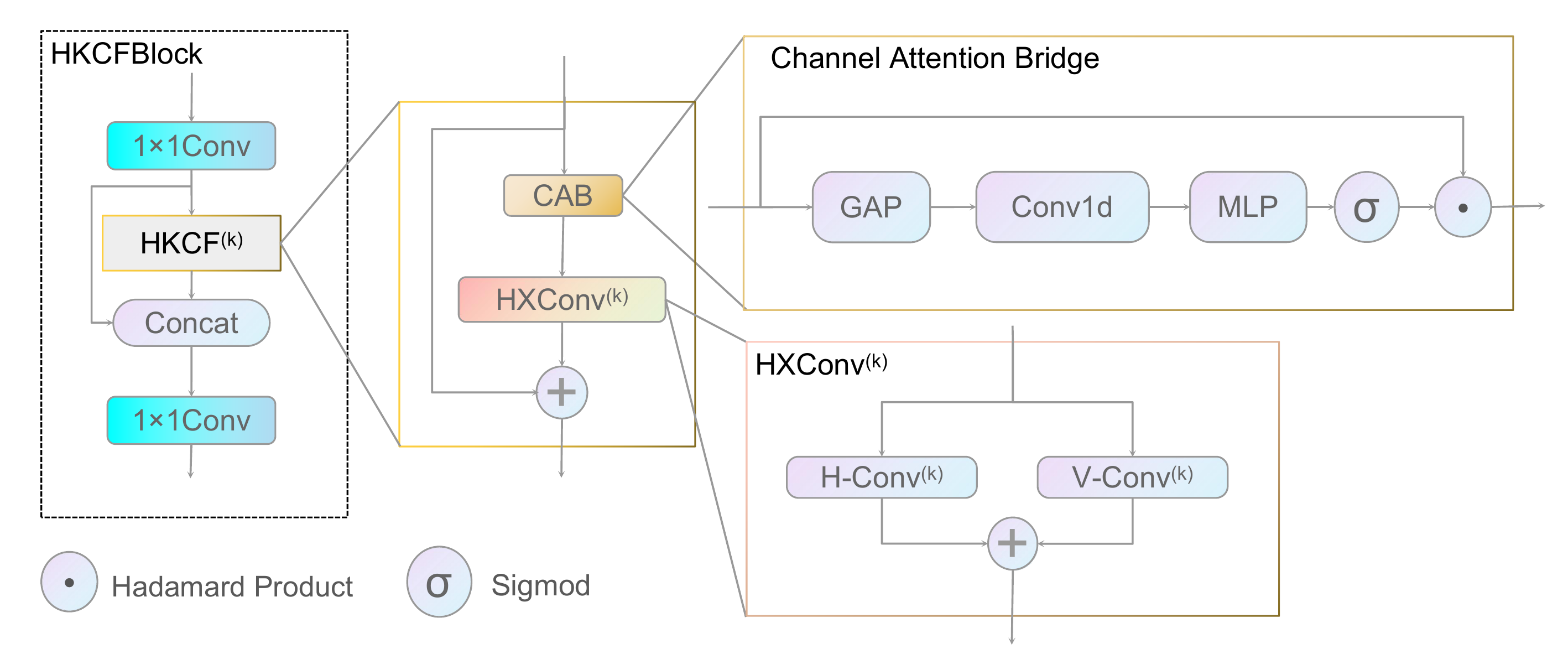}
  \caption{The internal structure of HKCF primarily consists of bottleneck structures and cross-convolutions. CBA is used for dimensionality reduction loss in dense bottleneck structures, and the selection of cross-convolution kernels follows HKSP.}
  \label{fig:hkxf}
\end{figure}

In addition to dense objective distribution, table cell segmentation also faces challenges due to extreme scale diversity caused by merged cells. Horizontal merging creates wide-spanning objects across multiple columns, while vertical merging generates tall and narrow objects spanning multiple rows. Such objectives demand multi-granularity perception capabilities from the model. The YOLO series addresses scale variation through multi-scale detection heads. High-resolution feature maps detect small objectives, while low-resolution ones focus on  large objectives. However, traditional fixed-size convolutional kernels struggle to adapt to the morphological diversity of cell features in tables. Inspired by YOLO-MS \cite{hksp}, which proposes matching objective diversity through kernel diversity and introduces the Heterogeneous Kernel Selection Protocol (HKSP) \cite{hksp}, we incorporate the HKSP concept and integrate asymmetric cross-convolution \cite{crossconv} to design a Heterogeneous Kernel Cross-Fusion (HKCF) module. As shown in Figure \ref{fig:hkxf}, the module uses a bottleneck structure to reduce computational complexity. As shown in Figure \ref{fig:hkxf}, the module uses a bottleneck structure to reduce computational complexity. The input feature map $I_{in}$ is first channel-reduced using $1 \times 1$ convolution to obtain low-dimensional features $F_{in}$. The $HKCF^{(k)}$ module with kernel size $k$ then extracts features $F_{out}$ in this reduced channel space. Finally, $F_{out}$ and $F_{in}$ are concatenated along the channel dimension. The original channel dimension is restored via another $1 \times 1$ convolution to output the feature map $I_{out}$. The computational flow can be formulated as:
\begin{equation}
  \begin{gathered}
    F_{in} =Conv(I_{in})
    \\
    F_{out} =HKCF^{(k)}(F_{in})
    \\
    I_{out} = Conv(Cat(F_{in}, F_{out}))
  \end{gathered}
  \label{eq:hkxfblock}
\end{equation}
As shown in Equation \ref{eq:hkxf}, within the Heterogeneous Kernel Cross-Fusion module $HKCF^{(k)}$, the input feature map $F_{in}$ first undergoes processing through the Channel Attention Bridge\cite{cba} (CAB). The CAB uses global average pooling and a multilayer perceptron to generate channel-wise weights, dynamically enhancing feature channels strongly correlated with object morphology, thereby alleviating information loss caused by the bottleneck structure's dimensionality reduction. To address the extreme narrow-tall and wide-flat challenges posed by merged cells, the module replaces standard convolution with heterogeneous cross convolution $HKCF^{(k)}$. This operation deploys horizontally extended kernels ($1 \times k$) and vertically extended kernels ($k \times 1$) in parallel. These dual branches respectively capture horizontal continuity features for column-spanning merged cells and vertical long-range dependencies for row-spanning merged cells. The outputs of these two branches sum to precisely adapt to the geometric characteristics of objects at varying scales. Following HKSP, cross-convolutions with kernel sizes {3, 5, 7} are progressively applied from shallow to deep feature fusion maps. Finally, the original input $F_{in}$ is added to the output feature map from $HXConv^{(k)}$ by residual connection, producing the final output feature map $F_{out}$.
\begin{equation}
    \begin{gathered}
        {F}_{out} = HXConv^{(k)}({CAB}({F}_{in})) + {F}_{in} \\
        HXConv^{(k)}(I) = HConv^{(k)}(I) + VConv^{(k)}(I)
    \end{gathered}
    \label{eq:hkxf}
\end{equation}%

\subsection{Loss function optimization}
In anchor-based instance segmentation frameworks, the object loss $\mathcal{L}_{obj}$ typically depends on Intersection over Union (IoU) and its variants to measure geometric deviations between predicted and ground-truth bounding boxes. However, the conventional CIoU loss \cite{ciou} has ambiguous optimization directions for objects with extreme aspect ratios because it couples aspect ratio and center distance computations. This results in reduced bounding box regression accuracy. To address this issue, we replace CIoU with the EIoU loss \cite{eiou}. EIoU explicitly decouples width-height optimization paths, enabling objected gradient direction adjustments for objects sensitive to aspect ratios. 

For mask loss $\mathcal{L}_{mask}$,the YOLO framework uses the binary cross-entropy loss ($\mathcal{L}_{BCE}$) as follows:

\begin{equation}
  \footnotesize
  \begin{aligned}
    \mathcal{L}_{BCE} = -\frac{1}{N} \sum_{i=1}^{N} \left[ \mathbf{M}_{{gt}}^{(i)} \log \mathbf{M}_{{pred}}^{(i)} + (1-\mathbf{M}_{{gt}}^{(i)}) \log (1-\mathbf{M}_{{pred}}^{(i)}) \right]
  \end{aligned}
  \label{eq:bce}
\end{equation}

\noindent where $N$ denotes the total number of pixels in the mask, $\mathbf{M}^{i}_{\text{gt}}$ represents the i-th pixel in the true mask, and $\mathbf{M}^{i}_{\text{pred}}$ is the i-th pixel in the predicted mask.

However, relying solely on the binary cross-entropy loss $\mathcal{L}_{BCE}$ neglects the inherent similarity between cell backgrounds and the overall table background, where background pixels dominate in quantity. To prevent the model from biasing towards the high-frequency background class, we integrate Dice Loss\cite{dice} into the base segmentation loss $\mathcal{L}_{base}$. Unlike cross-entropy loss, which focuses on per-pixel probability calibration, Dice Loss optimizes for region overlap, prioritizing the structural integrity of segmented objects. This approach enhances edge alignment and regional continuity. The Dice Loss formula is defined in Equation as follows:
\begin{equation}
  \mathcal{L}_{\text{Dice}(x, y)} = 1 - \frac{2 \sum \mathbf{M}_{\text{pred}}(x,y) \odot \mathbf{M}_{\text{gt}}(x, y)} {\sum \mathbf{M}_{\text{pred}}(x,y) + \sum \mathbf{M}_{\text{gt}}(x,y)}
  \label{eq:dice}
\end{equation}
\noindent where $\odot$ denotes element-wise multiplication.

The base loss $\mathcal{L}_{base}$ combines binary cross-entropy and Dice loss through summation:
\begin{equation}
\mathcal{L}_{\text{base}} = \mathcal{L}_{\text{BCE}} + \mathcal{L}_{\text{Dice}}.
\end{equation}
In the original loss function, YOLO normalizes the loss to balance the influence of objects of different sizes through inverse of the object's area term as follows:  
\begin{equation}
    \mathcal{L}_{{norm}} = \frac{1}{A}  \sum_{(x,y) \in {crop}(\Omega)} \left[\frac{1}{H' W'} \mathcal{L}_{\text{BEC}}(x,y) \right]
    \label{eq:norm_bce}
\end{equation}  
where $H'$ and $W'$ represent the height and width of the bounding-box cropped region $crop(\Omega)$ for the instance, and $A$ corresponds to the area (pixel count) of the object instance mask for normalization.  

This design aims to balance loss magnitudes across objects of different scales, preventing larger objects from dominating the optimization direction due to their pixel advantage. However, this inverse-proportional compensation mechanism exhibits inherent mathematical limitations: while the $1/A$ function approaches infinity as $A\rightarrow0^+$, aligning with the intuitive need for stronger compensation of small objects, its second derivative $\frac{d^2}{dA^2}(1/A)=2/A^3$ reveals rapidly increasing curvature characteristics. This mathematical property causes compensation intensity to exhibit near-vertical growth trends when target area falls below a critical threshold (e.g., $A<0.1$), potentially inducing gradient mutation issues.  

To establish a smoother scale adaptation mechanism, we introduce a logarithmic weighting function $W_s=1+\log(1/A)$ building upon the original normalization. The mathematical superiority of this design manifests in its differential properties: its first derivative $\frac{dW_s}{dA}=-1/A$ responds more gradually to small-scale variations compared to $\frac{d}{dA}(1/A)=-1/A^2$ of the original function. This moderation proves particularly significant in the critical small-target regime ($A<0.1$). Simultaneously, its second derivative $\frac{d^2W_s}{dA^2}=1/A^2$ maintains convexity while demonstrating substantially lower curvature than the $\frac{1}{A}$ function ($2/A^3$), thereby avoiding excessively aggressive compensation.  

From a function space perspective, these components form a complementary optimization: $\frac{1}{A}$ provides foundational compensation intensity ensuring necessary loss amplification for small objects, while $\log(1/A)$ acts as a modulator suppressing curvature mutation risks through its progressive growth characteristics. This dual composite structure maintains coherent limiting behavior—both functions approach infinity as $A\rightarrow0^+$, while $W_s$ converges to 1 as $A\rightarrow1^-$ ensuring larger objects experience no additional suppression. The resulting scale-aware loss function:  
\begin{equation}
\footnotesize
\begin{aligned}
    \mathcal{L}_{{scale-aware}} = \frac{1}{n} \sum_{i=1}^{n} \left[ \left( 1+\log \frac{1}{A_i} \right) \frac{1}{A_i} \sum_{(x,y) \in {crop}(\Omega)} \left[\frac{1}{H_i' W_i'} \mathcal{L}_{{base}}(x,y) \right] \right]
\end{aligned}
\label{eq:scale-aware}
\end{equation}  
where $n$ denotes the number of instances in the image, $H_i'$ and $W_i'$ represent the height and width of the bounding-box cropped region $crop(\Omega)$ for instance $i$, and $A_i$ corresponds to the area of instance $i$.  

The scale-aware loss achieves three-tiered optimization: at the microscopic level, the logarithmic term smooths gradient variations in the small-target regime; at the mesoscopic level, dual-function synergy maintains compensation continuity across transitional regions ($0.1\leq A_i\leq0.5$); at the macroscopic level, optimization stability for larger targets ($A_i>0.5$) is preserved. 

\subsection{Post-process optimization}

Anchor-based detection models typically rely on Non-Maximum Suppression (NMS) as a core post-processing operation to filter redundant detection boxes using an Intersection over Union (IoU) threshold. Traditional NMS performs robustly in sparsely distributed object scenarios but exhibits inherent limitations in dense deformed table cell detection: when adjacent cells exhibit highly overlapping bounding boxes due to geometric deformation, the IoU criterion mistakenly identifies them as the same instance, thereby suppressing correctly localized boxes with lower confidence scores. Although Soft-NMS \cite{soft-nms} partially alleviates over-suppression through a confidence decay mechanism, severe overlaps caused by complex object shapes still lead to inflated IoU values, failing to address the root cause. To resolve this, inspired by the mask competition strategy in SOLOv2 \cite{solov2}, we abandon the bounding-box suppression paradigm and introduce Mask-Driven Non-Maximum Suppression, which employs pixel-level Mask\_IoU for redundancy elimination. The metric is defined as:

\begin{equation}
  \text{Mask\_IoU} = \frac{| M_i \cap M_i |}{| M_i \cup M_j |}
  \label{eq:mask-iou}
\end{equation}

where $M_i$ and $M_i$ are binary matrices of two predicted masks. A low-confidence mask is suppressed only if its Mask\_IoU with a higher-confidence mask exceeds the threshold. Compared to the traditional IoU criterion, Mask\_IoU directly quantifies mask overlap, circumventing spatial mismatches induced by bounding box overlaps, thereby preserving more precise cell instances in complex table layouts.

\section{Experiments}
\subsection{Experimental settings}
The datasets involved in the experiments are DWTAL-s and DWTAL-l, both of which are experimented on an RTX 3090 with 24 GB of video memory, and the whole experiment relies on Python 3.8.19, PyTorch 1.13.0, and CUDA version 12.4.

For ablation studies, we don't use pretrained weights. Input images were resized to $640 \times 640$, with a batch size of 2. Training spanned 200 epochs using a Stochastic Gradient Descent (SGD) optimizer configured with momentum factor of 0.9, initial learning rate of 0.001, and weight decay of 0.0005.

In order to facilitate comparison experiments, all non-YOLO-based models were implemented using the MMDetection3 framework \cite{mmdet3} on the 24GB RTX 3090 GPU. ResNet-101 backbones were initialized with pretrained weights from Microsoft Research Asia (MSRA) \cite{mra}, trained on the ImageNet dataset \cite{imagenet}. For DWTAL-s, input images were set to $640 \times 640$ with a batch size of 2. Due to limited training resources, DWTAL-l experiments also used $640 \times 640$ inputs but with a reduced batch size of 1. Both datasets underwent 100 epochs of fine-tuning with the SGD optimizer, configured with momentum of 0.9, initial learning rate of 0.001, and weight decay of 0.0001.
\subsection{Evaluation metrics}
To comprehensively evaluate the performance of instance segmentation models, we adopt general metrics including mean Average Precision (mAP), model parameter count, and GFLOPs as quantitative criteria. mAP measures the model’s segmentation and classification capabilities by calculating average precision across multiple confidence thresholds. Specifically, precision P(t) and recall R(t) at a given confidence threshold t are first computed using Equation \ref{eq:PR}:

\begin{equation}
  \text{P(t)} = \frac{\text{TP(t)}}{\text{TP(t)} + \text{FP(t)}}, \quad \text{R(t)} = \frac{\text{TP(t)}}{\text{TP(t)} + \text{FN(t)}}
  \label{eq:PR}
\end{equation}

\noindent where FP(t) denotes false positives at confidence threshold t, and FN(t) represents false negatives at the same threshold.

Average Precision (AP) is defined as the area under the precision-recall (PR) curve. The per-class average precision AP(t) at confidence threshold t is calculated via Equation \ref{eq:AP}:
\begin{equation}
  \text{AP(t)} = \frac{1}{101} \sum_{r \in \{0, 0.01, 0.02, \dots, 1\}} \text{P(t)}_{\text{interp}}(r)
  \label{eq:AP}
\end{equation}
\noindent where \(\text{P(t)}_{\text{interp}}(r)\) denotes the maximum precision when the recall $R(t) \geqslant $ \(r\).

Finally, the mean Average Precision mAP(t) is obtained by averaging the AP(t) values across all classes at confidence threshold t using Equation \ref{eq:mAP}. Different confidence thresholds lead to distinct metrics: mAP@50 uses a confidence threshold of 0.5, reflecting baseline performance under lenient localization requirements, while mAP@50:95 rigorously evaluates the model’s robustness to object boundaries by averaging results across confidence thresholds from 0.5 to 0.95 with a step size of 0.05.

\begin{equation}
  \text{mAP}(t) = \frac{1}{N_{\text{classes}}} \sum_{i=1}^{N_{\text{classes}}} \text{AP}_i(t)
  \label{eq:mAP}
\end{equation}

The number of parameters reflects the model's complexity and storage requirements. Excessive parameters may lead to overfitting and deployment challenges, while insufficient parameters can limit feature representation capacity. GFLOPs quantifies the computational demand per inference in billions of floating-point operations, measuring computational efficiency. Models with high GFLOPs rely on high-performance GPUs and struggle to meet real-time requirements, whereas low-GFLOPs designs are suitable for real-time video processing but require architectural optimization to balance accuracy trade-offs. All GFLOPs values in this study are reported under the input size of $640 \times 640$.

\subsection{Quantitative result}
To validate the model’s effectiveness, we compares the proposed model with mainstream segmentation models on the DWTAL-s dataset, including two-stage models (Mask R-CNN \cite{mrcnn}, Cascade Mask R-CNN \cite{cmrcnn}), single-stage models (SOLOv2 \cite{solov2}, YOLACT \cite{yolact}), Transformer-based architectures (Mask2Former \cite{m2f}), and classical YOLO models (YOLOv5l-seg \cite{yolov5}, YOLOv8l-seg \cite{yolov8}, YOLOv11l-seg \cite{yolov11}). 

The experimental results on the DWTAL-s dataset, as shown in Table \ref{tab:compare-DWTAL-s}, demonstrate that the proposed OG-HFYOLO model achieves a mAP@50:95 of 74.23\%, outperforming mainstream two-stage instance segmentation models such as Mask R-CNN (62.5\%) and Cascade Mask R-CNN (62.1\%). It surpasses classical YOLOv8 (57.5\%) and the more advanced YOLOv11 (57.8\%) by 16.73\% and 16.43\%, respectively, and exceeds the highest-precision YOLOv5 variant (71.96\%) by 2.27\%. Additionally, the proposed model outperforms the Transformer-based Mask2Former (63.3\%) by 10.93\%. Regarding model parameter count, the introduction of a heterogeneous kernel cross-fusion architecture with relatively large convolutional kernels slightly increases the parameter size, though this remains manageable under current hardware storage constraints. In terms of inference speed, YOLACT has seriously lost model accuracy in the pursuit of extreme speed and model size, while the OG-HFYOLO model balances accuracy and speed, retaining the speed that a single-stage model should have. It surpasses both mainstream two-stage models and some single-stage models like SOLOv2 and Mask2Former in speed.
\begin{table}[!htbp]
  \centering
  \caption{The different model performance on DWTAL-s. The best results are in bold. }
  \resizebox{\linewidth}{!}{ 
  \begin{tabular}{c c c c c c c c}
      \toprule
      \multirow{2}{*}{Model} & \multirow{2}{*}{Backbone} & \multirow{2}{*}{Para (M)} & \multirow{2}{*}{GFLOPs} & \multicolumn{2}{c}{Bbox} & \multicolumn{2}{c}{Mask} \\
      \cmidrule(lr){5-6} \cmidrule(lr){7-8}
      & & & &   mAP@50 & mAP@50:95 & mAP@50 & mAP@50:95 \\
      \midrule  
      Mask RCNN\cite{mrcnn} & ResNet-101 & 62.963 & 245 & 0.752 & 0.671 & 0.736  & 0.625 \\
      Cascade Mask RCNN\cite{cmrcnn} & ResNet-101 & 96.013 & 978 & 0.752 & 0.679 & 0.741 & 0.621 \\
      Mask2Former\cite{m2f} & ResNet-101 & 62.995 & 230 & 0.742 & 0.656 & 0.742 & 0.644 \\
      SOLOv2\cite{solov2} & ResNet-101 & 65.221 & 243 & -&- & 0.637 & 0.451 \\
      YOLACT\cite{yolact}  & ResNet-101 & 53.719 & 84.719 &0.692 & 0.586  & 0.678 & 0.482 \\ 
      YOLOv5l-seg\cite{yolov5} & CSPDark & 92.27 & 147.5 & 0.9754 & 0.9144 & 0.9665 & 0.7196 \\
      YOLOv8l-seg\cite{yolov8} & CSPDark & 88 & 220.1 & 0.721 & 0.684 & 0.721 & 0.575 \\
      YOLOv11l-seg\cite{yolov11} & CSPDark & 53.22 & 141.9 & 0.721 & 0.688 & 0.721 & 0.578 \\
      \midrule
      ours & $CSPDark^{*}$ & 125.39 & 170 & \textbf{0.9896} &\textbf{ 0.9156} & \textbf{0.9851} & \textbf{0.7423} \\
      \bottomrule
  \end{tabular}
  \label{tab:compare-DWTAL-s}
  }
  \begin{tablenotes}     
  \scriptsize
      \item *:$CSPDark^{*}$ is based on the normal CSPDark backbone and introduces the GOE module. The overall backbone structure is the same as that of CSPDark.
   \end{tablenotes}
\end{table}

As shown in Table \ref{tab:compare-DWTAL-l}, the proposed model also achieves state-of-the-art segmentation accuracy metrics on the DWTAL-l dataset compared to other mainstream models, demonstrating its generalization capability across diverse datasets.

\begin{table}[!htbp]
  \centering
  \caption{The different model performance on DWTAL-l. The best results are in bold. }
  \resizebox{\linewidth}{!}{ 
  \begin{tabular}{c c c c c c c c}
      \toprule
      \multirow{2}{*}{Model} & \multirow{2}{*}{Backbone} & \multicolumn{2}{c}{Bbox} & \multicolumn{2}{c}{Mask} \\
      \cmidrule(lr){3-4} \cmidrule(lr){5-6}
      & &  mAP@50 & mAP@50:95 & mAP@50 & mAP@50:95 \\
      \midrule  
      Mask RCNN\cite{mrcnn} & ResNet-101 & 0.524 & 0.479 & 0.509 &  0.447\\
      Cascade Mask RCNN\cite{cmrcnn} & ResNet-101 & 0.524 & 0.488 & 0.521 & 0.451\\
      SOLOv2\cite{solov2} & ResNet-101 & - & - & 0.48 & 0.39 \\
      YOLACT\cite{yolact}  & ResNet-101 &0.446 & 0.308 & 0.401 & 0.225 \\ 
      YOLOv5l-seg\cite{yolov5}   & CSPDark & 0.860 & \textbf{0.803} & 0.8338 & 0.6134 \\
      YOLOv8l-seg\cite{yolov8} & CSPDark & 0.663 & 0.627 & 0.654 & 0.514 \\
      YOLOv11l-seg\cite{yolov11} & CSPDark & 0.664 & 0.633 & 0.654 & 0.521 \\
      \midrule
      ours & $CSPDark^{*}$ & \textbf{0.8685} & 0.8017 & \textbf{0.8457} & \textbf{0.6238} \\
      \bottomrule
  \end{tabular}
  }
  \label{tab:compare-DWTAL-l}
\end{table}

\subsection{Qualitative analysis}
\begin{figure}
  \centering
  \includegraphics[width=\linewidth]{image/compare.pdf}
  \caption{Comparison of the effects of different models:Subfigures (a)\textendash (c) are derived from the simpler DWTAL-s dataset, and (d)\textendash (f) originate from the more complex DWTAL-l dataset.}
  \label{fig:comp}
\end{figure}
Figure \ref{fig:comp} presents a comparative visualization of segmentation results across different models. Observations reveal that even seemingly straightforward cases like (a) exhibit severe missed detections in most mainstream models, whereas the proposed model achieves performance comparable to the advanced Mask2Former on this example and outperforms it across all subsequent test images. Among anchor-based models, YOLOv5 demonstrates the second-lowest missed detection rate after our proposed method. However, in scenarios like (d), the bounding box-based non-maximum suppression (NMS) not only suppresses valid cells but also introduces erroneous overlapping detections, as seen in (b) and (e). For images with blurred contours (f) and multi-colored background cells (e), the proposed model leverages its GOE module to achieve optimal detection results. In scale-variant cell detection tasks, such as the elongated cell in the bottom row of (b), SOLOv2 fails to accurately capture cell dimensions, while our model maintains robust performance.

\begin{figure}[!htbp]
  \centering
  \includegraphics[width=\linewidth]{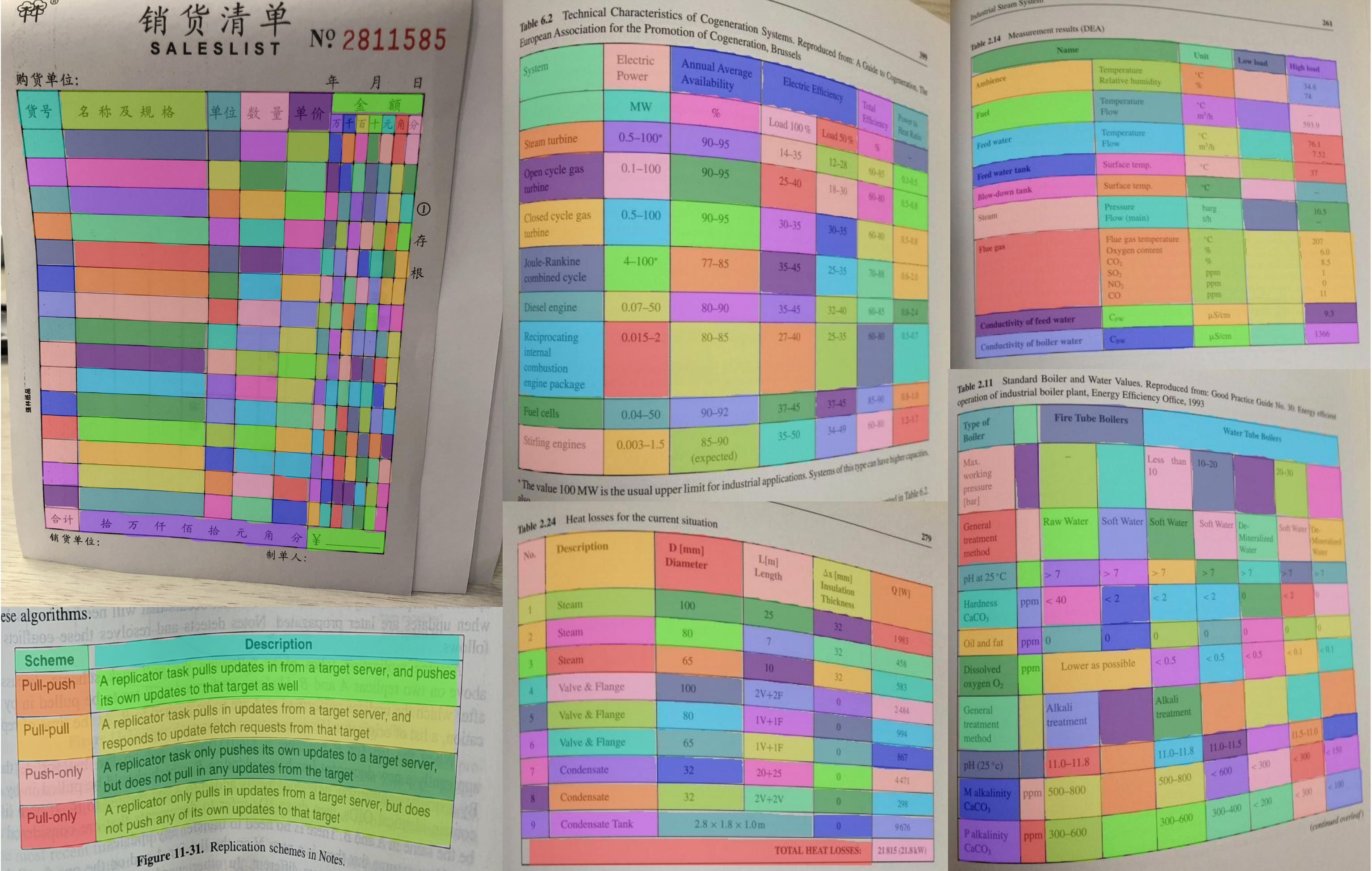}
  \caption{Our model result in real-world and other dataset. }
  \label{fig:gene}
\end{figure}

To further demonstrate the generalization capability of the proposed model for table cell segmentation, Figure \ref{fig:gene} showcases segmentation results on real-world photography scenarios and images from the Camcap dataset \cite{camcap}, using a model trained exclusively on DWTAL-l data. The left column displays natural scene photographs: the top-left image originates from real tables in the WTW dataset, and the bottom-left represents actual captured scenes. The right four subfigures correspond to Camcap samples. From the results, the OG-HFYOLO model can still achieve good segmentation results, which is enough to prove that the model has strong generalization ability.

\subsection{Ablation study}
\subsubsection{Impact of Anchor Mechanism}
With the advancement of deep learning, detection-oriented models have progressively transitioned from anchor-based to anchor-free \cite{anchorfree} mechanism to pursue higher speed and smaller parameter footprints. Starting from YOLOv5, models began reducing reliance on predefined anchors, and by YOLOv8, anchor-free mechanism became the standard. Comparative experiments clearly demonstrate that YOLOv8l-seg and YOLOv11l-seg, which adopt anchor-free mechanism, exhibit significantly lower accuracy than their YOLOv5l-seg counterparts within the same YOLO series. To validate the superiority of anchor-based mechanism for the current task, ablation studies on anchor mechanisms were conducted. As shown in Table \ref{tab:anchor-mode}, across both derived datasets, all metrics for both YOLOv5 and OG-HFYOLO models achieve over 10\% significant improvement when using anchor-based mechanism, highlighting anchor-based effectiveness in this research context.
\begin{table}[!htbp]
  \centering
  \caption{Impact of Anchor Mechanism. The best results are in bold. }
  \resizebox{\linewidth}{!}{ 
  \begin{tabular}{c c c c c c}
      \toprule
      \multirow{2}{*}{Dataset} & \multirow{2}{*}{mode} & \multicolumn{2}{c}{Bbox} & \multicolumn{2}{c}{Mask} \\
      \cmidrule(lr){3-4} \cmidrule(lr){5-6}
      & &   mAP@50 & mAP@50:95 & mAP@50 & mAP@50:95 \\
      \midrule
      \multicolumn{6}{c}{YOLOv5} \\
      \midrule
      \multirow{2}{*}{DWTAL-s} & anchor-free & 0.72 & 0.682 & 0.72 & 0.573 \\
      & anchor-base & \textbf{0.9754} & \textbf{0.9144} & \textbf{0.9665} & \textbf{0.7196} \\
      \midrule
      \multirow{2}{*}{DWTAL-l} & anchor-free & 0.664 & 0.629 & 0.654 & 0.515 \\
      & anchor-base & \textbf{0.860} & \textbf{0.803} & \textbf{0.8338} & \textbf{0.6134} \\
      \midrule
      \multicolumn{6}{c}{OG-HFYOLO} \\
      \midrule
      \multirow{2}{*}{DWTAL-s} & anchor-free & 0.721 & 0.68 & 0.621 & 0.318 \\
      & anchor-base & \textbf{0.9896} & \textbf{0.9156} & \textbf{0.9851} & \textbf{0.7423}  \\
      \midrule
      \multirow{2}{*}{DWTAL-l} & anchor-free & 0.662 & 0.619 & 0.563 & 0.3 \\
      & anchor-base & \textbf{0.8685} & \textbf{0.8017} & \textbf{0.8457} & \textbf{0.6238}  \\
      \bottomrule
  \end{tabular}
  \label{tab:anchor-mode}
  }
\end{table}

\subsubsection{Effect of the proposed method}
Table \ref{tab:abystudy} presents the ablation study results of the proposed method on the DWTAL-s dataset. The Gradient Orientation-aware Extractor (GOE) is designed to capture richer texture information and mitigate detection challenges caused by dense object distributions. The Heterogeneous Kernel Cross Fusion (HKCF) and scale-aware Loss address severe scale and aspect ratio variations, while the MASK-NMS algorithm optimizes post-processing to handle complex shapes and crowded instances. The challenges inherent in the derived dataset are interdependent, meaning solving isolated issues yields limited performance gains. For instance, while texture extraction by GOE partially addresses detection difficulties, it fails to resolve scale variation problems, resulting in marginal improvements (e.g., only a 0.44\% increase in mask mAP@50:95 when GOE is introduced alone). Similarly, isolated integration of HKCF and scale-aware Loss improves mask mAP@50:95 by merely 0.09\% and 0.48\%, respectively. However, as demonstrated in Table \ref{tab:abystudy}, synergistic integration of complementary modules yields significant enhancements. For example, combining GOE with scale-aware Loss boosts mask mAP@50:95 by 1.29\%. These results confirm that while each proposed module is individually effective, their collaborative integration is essential to achieve optimal performance.

\begin{table}[!htbp]
  \centering
  \caption{Effect of the proposed method. The best results are in bold. }
  \resizebox{\linewidth}{!}{ 
  \begin{tabular}{c c c c c c c c c c}
      \toprule
      \multirow{2}{*}{GOE} & \multirow{2}{*}{HKCF} & \multirow{2}{*}{Scale-Aware Loss} & \multirow{2}{*}{MASK-NMS} & \multicolumn{2}{c}{Bbox} & \multicolumn{2}{c}{Mask} & \multirow{2}{*}{Para (M)} & \multirow{2}{*}{GFLOPs} \\
      \cmidrule(lr){5-6} \cmidrule(lr){7-8}
      & &  & & mAP@50 & mAP@50:95 & mAP@50 & mAP@50:95 \\
      \midrule
      & & & &  0.9754 & 0.9144 & 0.9665 & 0.7196 & 92.27 & 147.5 \\
      \midrule
      $\surd $  & & & & 0.9755 & 0.9106 & 0.9657 & 0.724 & 104 & 152 \\
      & $\surd $  & &  & 0.9753 & 0.9122 & 0.9645 & 0.7205 & 113.66 & 168\\
      & & $\surd $  &  & 0.9766 & 0.9087 & 0.9715 & 0.7244 & 92.27 & 147.5 \\
      \midrule
      $\surd $  & & $\surd $ &  & 0.9774 & 0.9022 & 0.9724 & 0.7325 & 104 & 152 \\
      & $\surd $ & $\surd $  &  & 0.9775 & 0.9037 & 0.9728 & 0.7318 & 113.66 & 168 \\
      \midrule
      $\surd $ & $\surd $  & $\surd $ & & 0.9778 & 0.903 & 0.9731 & 0.7329 & 125.39 & 170 \\
      $\surd $ & $\surd $  & $\surd $  & $\surd $  & \textbf{0.9896} & \textbf{0.9156} & \textbf{0.9851} & \textbf{0.7423} & 125.39 & 170 \\
      \bottomrule
  \end{tabular}
  \label{tab:abystudy}
  }
\end{table}

\subsection{Effect of module in the backbone}
To clarify why the proposed model does not adopt the more advanced backbone architecture from YOLOv11, which introduces the flexible C3k2 module \cite{yolov11} for feature extraction and integrates a C2PSA attention mechanism \cite{yolov11} at the final stage to consolidate backbone features, ablation studies were conducted across different backbone configurations under identical experimental settings (including the same anchor mechanism, training strategy, and proposed modules). As demonstrated in Table \ref{tab:C3-C3k2}, the baseline C3 architecture achieves optimal performance on both datasets, outperforming variants incorporating YOLOv11's enhanced components. The experimental results demonstrate that optimal results can be achieved by using the C3 module directly.

\begin{table}[!htb]
  \centering
  \caption{Effect of different backbone. The best results are in bold. }
  \resizebox{\linewidth}{!}{ 
  \begin{tabular}{c c c c c c c c}
      \toprule
      \multirow{2}{*}{Dataset} & \multirow{2}{*}{Backbone} & \multicolumn{2}{c}{Bbox} & \multicolumn{2}{c}{Mask} \\
      \cmidrule(lr){3-4} \cmidrule(lr){5-6}
      & &   mAP@50 & mAP@50:95 & mAP@50 & mAP@50:95 \\
      \midrule
      \multirow{2}{*}{DWTAL-s} & YOLOv5-backbone &  \textbf{0.9896} &  \textbf{0.9156} &  \textbf{0.9851} & \textbf{0.7423} \\
      & YOLOv11-backbone & 0.978 & 0.9046 & 0.9733 & 0.7295 \\
      \midrule
      \multirow{2}{*}{DWTAL-l} & YOLOv5-backbone & \textbf{0.8685} & \textbf{0.8017} & \textbf{0.8457} & \textbf{0.6238} \\
      & YOLOv11-backbone & 0.8607 & 0.7889 & 0.8379 & 0.6139 \\
      \bottomrule
  \end{tabular}
  \label{tab:C3-C3k2}
  }
\end{table}

\section{Conclusion}
To address spatial coordinate localization in deformed table structures, we propose a novel instance segmentation framework. This approach establishes a novel pixel-level parsing paradigm that significantly outperforms traditional detection-based localization methods. Compared with traditional detection methods, we establish a new localization paradigm grounded in pixel-level parsing. To facilitate this research direction, we construct the DWTAL dataset specifically designed for deformed table cell localization tasks. Its fine-grained spatial coordinate annotations fill the critical benchmark data gap in this research domain. To address the challenges of dense object distributions and extreme scale variations in DWTAL dataset, we further propose the OG-HFYOLO model for accurate instance segmentation of deformed table cells. The model incorporates several key innovations: the Gradient Orientation-aware Extractor (GOE) enhances contour perception for densely packed objects; the Heterogeneous Kernel Cross-Fusion (HKCF) and scale-aware Loss address challenges posed by severe scale variation; and Mask-based Non-Maximum Suppression (MASK-NMS) prevents erroneous suppression caused by overlapping bounding boxes. 

This work leverages instance segmentation to acquire precise spatial coordinates of cells in deformed tables, a mid-to-upstream task in table structure recognition. Downstream tasks involving logical coordinate processing can adopt the operational framework of LGPMA, optimizing the workflow through systematic integration of Computer Graphics principles and geometric topology theory. Furthermore, challenges encountered in the derived dataset, such as dense object arrangements and scale diversity, are common in fields like medical cell segmentation and remote sensing image analysis. Consequently, the proposed methods provide constructive insights for addressing similar challenges in these fields.

\section*{Acknowledgments}
This work is supported by the Natural Science Foundation of China (No. 62366034).

\bibliographystyle{unsrt}  
\bibliography{references}

\end{document}